\documentclass[journal]{IEEEtran}

\usepackage{graphicx}
\graphicspath{{./figures/}}

\usepackage{amsmath}
\usepackage{bm}
\usepackage{amssymb}

\usepackage{soul}
\usepackage{color, colortbl}

\usepackage{subcaption}
\usepackage{booktabs}
\usepackage{placeins}
\usepackage{verbatim}

\definecolor{r1}{rgb}{0.0, 0.0, 0.0}

\usepackage{tikz}
\newcommand\copyrighttext{%
  \footnotesize \textcopyright 2018 IEEE. Personal use of this material is permitted.
  Permission from IEEE must be obtained for all other uses, in any current or future
  media, including reprinting/republishing this material for advertising or promotional
  purposes, creating new collective works, for resale or redistribution to servers or
  lists, or reuse of any copyrighted component of this work in other works.}
\newcommand\copyrightnotice{%
\begin{tikzpicture}[remember picture,overlay]
\node[anchor=south,yshift=0pt] at (current page.south) {\fbox{\parbox{\dimexpr\textwidth-\fboxsep-\fboxrule\relax}{\copyrighttext}}};
\end{tikzpicture}%
}

\begin{document}

%
% paper title
% Titles are generally capitalized except for words such as a, an, and, as,
% at, but, by, for, in, nor, of, on, or, the, to and up, which are usually
% not capitalized unless they are the first or last word of the title.
% Linebreaks \\ can be used within to get better formatting as desired.
% Do not put math or special symbols in the title.
%\title{Opportunistic Test-Time Feature Acquisition}
%\title{Cost-Aware Test-Time Feature Acquisition}
%\title{Context and Cost Aware Test-Time Feature Acquisition}
%\title{An Opportunistic Method for Cost-Aware Test-Time Feature Acquisition}
%\title{Denoising Autoencoders with Binary Representation Layers for Cost-Aware Feature Acquisition at Test-Time}
\title{Dynamic Feature Acquisition Using Denoising Autoencoders}

\author{{\normalsize Mohammad Kachuee, Sajad Darabi, Babak Moatamed, Majid Sarrafzadeh}\\ UCLA  Computer Science Department}

\author{Mohammad~Kachuee,~\IEEEmembership{Student Member,~IEEE,}
        Sajad~Darabi,~\IEEEmembership{Student~Member,~IEEE,}
        Babak~Moatamed,~\IEEEmembership{Student~Member,~IEEE,}
        and~Majid~Sarrafzadeh,~\IEEEmembership{Fellow,~IEEE}% <-this % stops a space
\thanks{Authors are with the UCLA Department of Computer Science. Address: UCLA Computer Science Dept., \#391 Eng. VI, 404 Westwood Plaza, CA 90095, USA (Email: mkachuee@cs.ucla.edu).
}% <-this % stops a space
%\thanks{J. Doe and J. Doe are with Anonymous University.}% <-this % stops a space
%\thanks{Manuscript received April 19, 2005; revised August 26, 2015.}}
}

% make the title area
\maketitle
\copyrightnotice

\markboth{Accepted for publication in IEEE Transactions on Neural Networks and Learning Systems (TNNLS)}{}

\begin{abstract}

%Objectives and goals\\
%Proposed ideas and methods\\
%Results\\
%Conclusion and Significance\\

In real-world scenarios, different features have different acquisition costs at test-time which necessitates cost-aware methods to optimize the cost and performance trade-off. This paper introduces a novel and scalable approach for cost-aware feature acquisition at test-time. The method incrementally asks for features based on the available context that are known feature values. The proposed method is based on sensitivity analysis in neural networks and density estimation using denoising autoencoders with binary representation layers. In the proposed architecture, a denoising autoencoder is used to handle unknown features (i.e., features that are yet to be acquired), and the sensitivity of predictions with respect to each unknown feature is used as a context-dependent measure of informativeness. We evaluated the proposed method on eight different real-world datasets as well as one synthesized dataset and compared its performance with several other approaches in the literature. According to the results, the suggested method is capable of efficiently acquiring features at test-time in a cost- and context-aware fashion.

%In real-world scenarios, different features have different acquisition costs at test-time which necessitates cost-aware methods to optimize the cost and performance trade-off. This paper introduces a novel and scalable approach for cost-aware feature acquisition at test-time. The proposed method is based on sensitivity analysis in neural networks and density estimation using denoising autoencoders with binary representation layers. Here, a denoising autoencoder is used to handle unknown features (i.e., features that are not acquired), while the sensitivity of predictions with respect to each unknown feature is used as a context-dependent measure of informativeness. We evaluated the proposed method on seven different real-world datasets as well as one synthesized dataset and compared its performance with other approaches in the literature. According to the results, the suggested method is capable of efficiently acquiring features at test-time in a cost- and context-aware fashion.

\end{abstract}

% Note that keywords are not normally used for peerreview papers.

%---------------------------------------------------------
\section{Introduction}
\label{sec:introduction}

%the main topic and its importance \\
% In the literature, predictability and scalability have always 
%  always been of great attention. However, recently, with the widespread use of machine learning in real-world applications and industrial setups, \hlgreen{Needs work!}

%Feature selection \\
%applications, examples \\
%high-level survey: static vs dynamic cost\\

%\hlgreen{FIXME: select and use a consistent acronym OPFA, DPFS, etc.}
\IEEEPARstart{F}{eature} selection methods have been largely studied in the literature. Usually, the main goal of feature selection is defined as selecting a subset of available features to increase the prediction performance and to reduce over-fitting. In real world scenarios, however, the cost of extracting or acquiring each feature is different from other features. The cost difference can be due to various factors such as differences in computational load in the extraction of features \cite{bay2008speeded,kachuee2017cuffless}, user disruptions in computer and user interactions \cite{early2016dynamic}, patient pain in medical procedures and tests \cite{sharpe1995dealing}, and so forth \cite{krishnapuram2011cost}. In these scenarios, selecting a feature that may only marginally contribute to an increase in the prediction accuracy which entails high costs would be unacceptable. In other words, there exists a trade-off between the feature cost and prediction performance that should be considered in the algorithm design.

To overcome this issue, there are methods suggested in the literature trying to adapt feature selection algorithms to consider the cost of each feature \cite{liu2017cost,ghasemzadeh2015power,min2014feature,cao2013optimized}. However, another point of concern that requires attention is that selecting a fixed set of features to be used during the training phase and using them at test-time would not be an optimal solution; as it neglects the potential interdependence between features. In many scenarios, there are features that are either freely available or easy to acquire at test-time. An optimal decision about other features to include in the analysis can be highly dependent on them. For instance, a doctor decides whether to prescribe an MRI scan based on the patient's current available information such as age, gender, symptoms and so on. In this example, having a fixed list of required tests and asking patients to provide the results of these tests for a clinical visit, would result in the high cost of MRI for all patients. In other words, the decision to include each feature should be based on the learned system dynamics as well as the available information at test-time.

%Main contributions of the paper list\\
%Paper organization \\
%In this paper, we address the problem of cost-aware and context-aware feature acquisition during the test phase. 

In this paper, we suggest a novel approach for feature acquisition considering costs at test-time (FACT). The proposed solution is capable of incrementally asking for features to be included in the prediction based on the current available context and user-defined feature costs. The rest of the paper is organized as follows. Section~\ref{sec:Related Work} briefly reviews the current relevant literature. Section~\ref{sec:Methodology} introduces the suggested approach including theoretical and implementation details. Section~\ref{sec:Experimental Results} presents the results of using the suggested method and compares them with the state-of-the-art approaches in the literature. Finally, Section~\ref{sec:Conclusion} concludes the paper.

%---------------------------------------------------------
\section{Related Work}
\label{sec:Related Work}
%context unaware, training-time \\
One of the approaches to incorporate feature acquisition costs or feature costs in general is considering the feature costs during the training phase and trading off the prediction accuracy with the prediction cost. An example of these approaches is limiting the number of features that are actually used in the predictor model by using $L_1$ regularization \cite{efron2004least}. In this method, the $L_1$ regularization enforces weights corresponding to certain features to be zero, and hence they can be omitted during the test phase. There are other methods in the literature that try to define and solve optimization problems over both the prediction performance and prediction costs \cite{greiner2002learning,ghasemzadeh2015power,ji2007cost}. Nevertheless, in all these methods, the final set of selected features is fixed and these methods fail to capture and take the advantage of the contextual information available at test-time.

%Simple feature selection based \\
%dynamic test-time feature selection: chapter 7.4 of (Classifier Cascades and Trees for Minimizing Feature Evaluation Cost) \\

One intuitive approach to incorporate feature costs during the training phase, while considering the available context during the test phase, is using the idea of decision trees. One of the most famous examples of this approach is the face detection cascade classifier by Viola and Jones \cite{viola2004robust}. While their goal was to increase the prediction speed by rejecting negative samples as soon as possible within a cascade of classifiers, many papers followed their architecture and incorporated feature cost in creating cascade predictors \cite{chen2012classifier,xu2012greedy}. One main drawback of cascade approaches is that cascades are only applicable to problems with a considerable class imbalance such as face detection or spam email detection. In these cases, the number of negative samples is significantly higher than the number of positive samples. However, there are many real-world applications in which the classes are relatively balanced such as document classification or image classification. To overcome this issue, in \cite{xu2014classifier,karayev2012timely} authors suggested the idea of classifier trees instead of classifier cascades to handle the problems where cascades are not applicable.

%Cascade based : works for imbalanced, feature groups should be hard-coded tree based and cascade-based are not strong in certain problems with high complexity, not instance specific and less flexible \\
While cascade and tree based test-time feature acquisition methods are shown to perform reasonably well in many scenarios; there are many problems and applications such as large-scale image classification, voice recognition, natural language processing, etc. where tree and cascade classifiers are not intrinsically strong enough to make accurate predictions. Another important limitation of cascade and tree based approaches is, while they include the context information to some extent, their feature query decisions are not truly instance specific. Specifically, they are limited by the fixed predetermined structure of the tree that enforces the features to be acquired at each tree node.

%Immitation learning :hard to train, Reinforcement learning based: , creating new models for the query rather than using the information captured in the model\\
In order to address these issues, recently, there has been great attention toward using learning methods to solve the generic problem of cost-sensitive and context-aware feature acquisition. He \textit{et al.} \cite{he2012cost} suggested a method based on imitation learning that trains a model that is able to predict an optimal feature query decision to be made given the available features. Contardo \textit{et al.} \cite{contardo2016recurrent,contardo2016sequential} introduced the idea of defining the problem as a reinforcement learning problem and solving it as a separate problem. While these methods are successful in terms of truly incorporating the test-time context information to the decisions, they require extra effort of training a feature query model in addition to the target predictor.

%Sensitivity based: Ubicomp 2015 paper\\
%Recently, with the advancement of strong machine learning algorithms that not only learn how to make predictions but also learn how
An alternative idea for measuring the informativeness of features given the context is using sensitivity analysis at test-time to measure the influence of each feature on the predictions. Early \textit{et al.} \cite{early2016test} introduced a method based on sensitivity analysis that exhaustively measures the impact of acquiring each feature on the prediction outcome. Their solution does not require training any other model, and it works in conjunction with almost any supervised learning algorithm. However, exhaustive sensitivity measurement is computationally expensive. It is impractical in problems with a large number of features to exhaustively examine the sensitivity with respect to each unknown feature. 

%Connection to our idea: one-shot, feature query is coupled with learning algorithm, scalable\\
In this paper we suggest a novel approach that is based on the idea of sensitivity analysis. The proposed approach incrementally asks for features based on the feature acquisition costs and the expected effect each feature can induce on the prediction. Furthermore, the devised method uses back-propagation of gradients and binary representation layers in neural networks to address the computational load as well as scalability concerns.
\textcolor{r1}{In an earlier work \cite{kachuee2017context}, we introduced the idea of sensitivity analysis as a method for dynamic feature selection. However, in this paper, we extend the idea by considering feature acquisition costs, introducing improvements such as feature encoding, and conducting more detailed experiments and analysis.}

%---------------------------------------------------------
\section{Proposed Method}
\label{sec:Methodology}

\subsection{Problem Definition}
In this paper, we consider the problem of predicting target classes ($\bm{y}\in R^r$) corresponding to a given feature vector ($\bm{x} \in R^d$). Each feature vector consists of known features as well as unknown features (i.e., missing values) that are set to zero. The complete feature vector without any missing values is denoted by $\tilde{\bm{x}}$. To indicate unknown features, a vector $\bm{k}\in\{0,1\}^d$ is defined that acts as a mask and indicates known and unknown features with one and zero values, respectively. In addition, we define a feature acquisition cost vector ($\bm{c} \in R^d$) that defines the cost of acquiring each feature.

%incremental prob def, feature query, cost, time step\\
For simplicity of analysis, we consider the incremental problem of having a feature vector ($\bm{x}^t$) and the corresponding mask vector ($\bm{k}^t$) at time step $t$. Additionally, we consider the cost values to be time dependent and defined for each time step ($\bm{c}^t$). Using this notation, at each time step $t$, the current feature vector can be represented as
\begin{equation}
    x^{(t)}_{j} = 
    \begin{cases}
                    0 & k^t_{j}=0 \\
                    \tilde{x_{j}} & k^t_{j}=1 \\
    \end{cases}
    ,
\end{equation}
which is acquired at the total cost of
\begin{equation}
    C_{total}^{t} = (\bm{k}^t - \bm{k}^0)^T \bm{c}^t.    
\end{equation}
Apart from this, at each time step, we have an expected prediction value ($\bm{y}^t$) using a predictor function ($h$) that takes $\bm{x}^t$ as input:
\begin{equation}
    \bm{y}^t = h(\bm{x}^t) = h(x_{1}^t, x_{2}^t, \dots, x_{d}^t) \;\;.
\end{equation}

In this setup, we define the feature query operator ($q$) as a function that acquires the value of feature $j$ in the incomplete feature vector $\bm{x}^t$ and outputs the feature vector of the next time step, $\bm{x}^{t+1}$:
\begin{equation}
\begin{aligned}
    \bm{x}^{t+1} = q(\bm{x}^t, j),
    \text{where ~~~~~~~~~~} \\ k_{j}^{t+1} - k_{j}^{t} = 1  \text{ and } k_{i}^{t+1} - k_{i}^{t} = 0 ~ (i \ne j) \;\; .
\end{aligned}
\end{equation}

%Target (with or without considering the feature acquisition cost?)\\
Furthermore, we define the desired feature to be queried at time $t$ as the feature that decreases the prediction error significantly, while at the same time, incurs a low acquisition cost. Mathematically speaking, we can use prediction accuracy improvement per acquisition cost as a measure of efficiency for the feature query. Accordingly, the desired feature to be queried at time step $t$ can be found by

\begin{equation}
    j^{t}_{sel} = \underset{j \in \{1 \dots d\}|k_j^t=0}{argmin} | \tilde{\bm{y}} - h(q(\bm{x}^t, j))| \; . \; c_j^t \;\; , 
\label{eq:opt}
\end{equation}
where $\tilde{\bm{y}}$ is the ground-truth target value, \textcolor{r1}{and $1$ is bias value in order to prevent the first term from becoming zero}. It is worth mentioning that the solution introduced here is basically an incremental solution that greedily selects features to be acquired at each step.

Table~\ref{tab:notations} presents a summary of the notations used throughout the paper.

\begin{table}%
\renewcommand{\arraystretch}{1.3}
\caption{The summary of the notations used throughout the paper.}
\label{tab:notations}
\begin{minipage}{\columnwidth}
\begin{center}
\resizebox{\columnwidth}{!}{
\begin{tabular}{ll}
\toprule

\textbf{Notation} & \textbf{Description} \\
\midrule
{\large$\tilde{\bm{y}} $} $ \in R^r$ & Ground-truth target values  \\
{\large$\bm{y} $} $\in R^r$ & Predicted target values  \\
{\large$\bm{x} $} $\in R^d$ & Incomplete feature vector at test-time \\
{\large$\tilde{\bm{x}}$ } $\in R^d$ & Complete feature vector without any missing values \\
{\large$\bm{x_{bin}}$} $\in R^{d \times l}$ & Binary representation of the feature vector \\
{\large$\bm{x'}$} $ \in R^{d}$ & Reconstructed feature vector \\
{\large $\bm{x'_{bin}} $}$ \in R^{d \times l}$ & Binary representation of the reconstructed feature vector \\
{\large$\bm{k} $} $\in\{0,1\}^d$ & Mask vector indicating known and unknown features \\
{\large$\bm{c} $} $ \in R^d$ & Feature acquisition cost vector\\
{\large$\bm{z} $} $\in R^{d'}$ & Encoded feature vector \\

\bottomrule
\end{tabular}
}
\end{center}
\end{minipage}
\end{table}%

\subsection{Sensitivity-based Feature Acquisition}
%Sensitivity based approach, feature query criterion\\
While (\ref{eq:opt}) suggests what features to be acquired at each step, directly using this equation is not practical. The reason behind this is the first term in this equation is usually not known and is difficult to estimate. To resolve this issue, it is possible to use the sensitivity of model predictions with respect to each missing feature as a measure for the potential impact of that feature on the final predictions.  As a result, (\ref{eq:opt}) can be rewritten using the suggested sensitivity measure as:
\begin{equation}
    j^{t}_{sel} = \underset{j \in \{1 \dots d\}|k_j^t=0}{argmax} \frac{Sensitivity(h(\bm{x}^t),j)}{c_j^t} \;\;.
\label{eq:opt_sense}
\end{equation}

Note that because a higher sensitivity is synonymous with a more informative feature to select, the argmin function in (\ref{eq:opt}) is replaced by argmax. Furthermore, the prediction sensitivity with respect to input j can be defined as
\begin{equation}
\begin{aligned}
    Sensitivity (h(\bm{x}^t),j) = \mathbb{E}_{x_j} (\frac{\partial h^t(\bm{x}^t)}{\partial x_{j}}) \\
    = \int |\frac{\partial h(\bm{x}^t)}{\partial x_{j}}|\: p(x_{j}|\bm{x}^t;h^t)  \: dx_{j} \;\;.
\end{aligned}
\end{equation}
In this equation, the first term corresponds to the derivative of the predictor function with respect to each missing feature. The second term is the confidence of inferring the $j$'th feature given the context and model parameters. By substituting this into (\ref{eq:opt_sense}), the feature query criterion can be written as
\begin{equation}
    j^{t}_{sel} = \underset{j \in \{1 \dots d\}|k_j^t=0}{argmax} \frac{\int |\frac{\partial h(\bm{x}^t)}{\partial x_{j}}|\: p(x_{j}|\bm{x}^t;h^t)  \: dx_{j}}{c_j^t} \;\;.
\label{eq:opt_deriv}
\end{equation}

Furthermore, the continuous integral in (\ref{eq:opt_deriv}) can be approximated by a discrete summation:
\begin{equation}
    j^{t}_{sel} = \underset{j \in \{1 \dots d\}|k_j^t=0}{argmax} \frac{\sum_{x_{j}\in RS} |\frac{\partial h(\bm{x}^t)}{\partial x_{j}}|\: p(x_{j}|\bm{x}^t;h^t)}{c_j^t} \;\; , 
\label{eq:opt_summation}
\end{equation}
where $RS$ is a set of samples from the range of possible values that can be taken by each feature. By adjusting the granularity of the values in the $RS$ set, one can trade-off between the approximation accuracy and the computational load of the expected value approximation.

\begin{figure*}
\centering
  \includegraphics[width=0.65\linewidth]{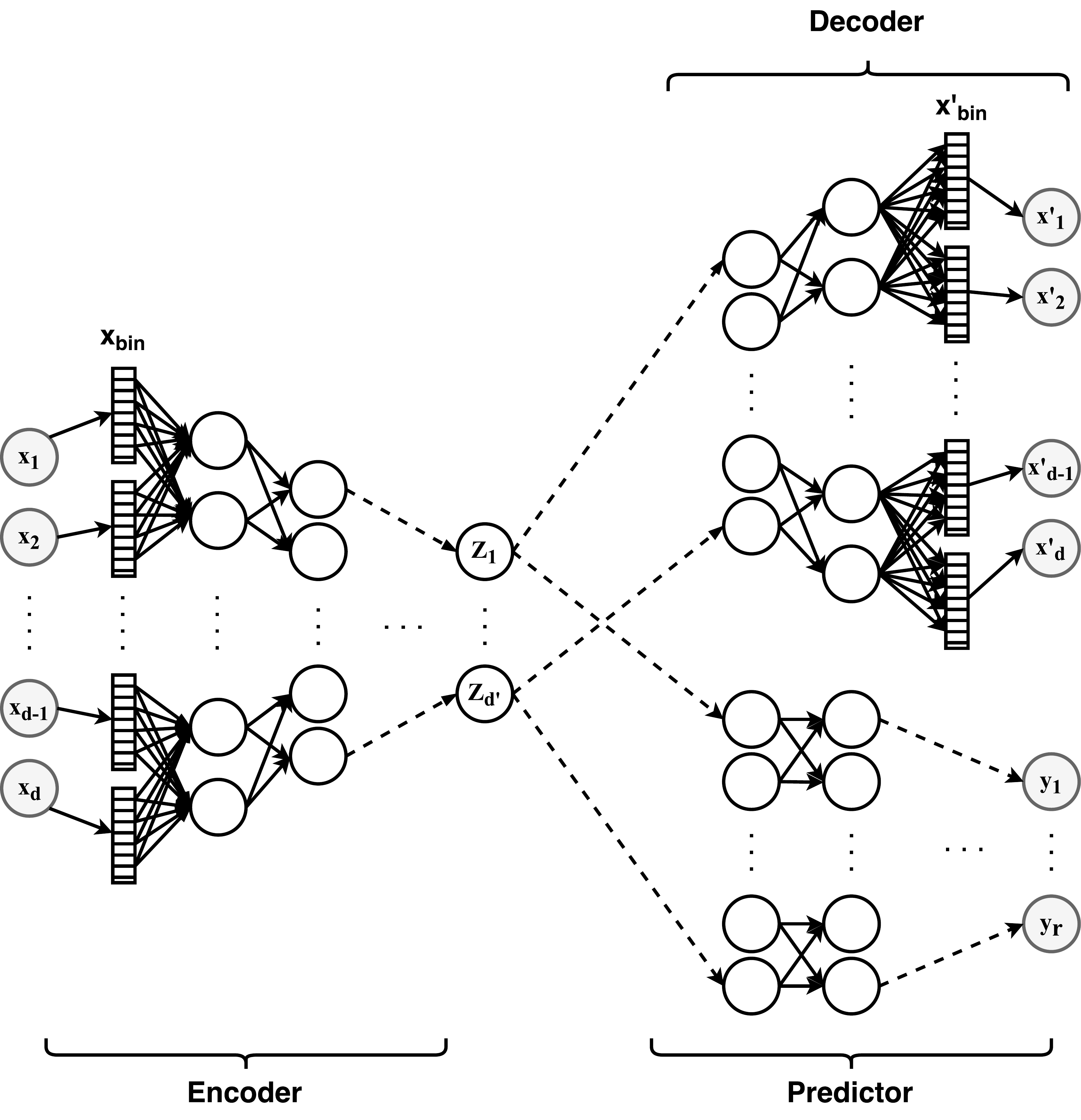}
  \caption{Network architecture of the proposed method including: encoder, decoder, and predictor parts. The encoder part is responsible for handling missing features. The decoder part is used for feature density estimation. The predictor is responsible for making predictions; additionally, its derivatives with respect to inputs are used for measuring sensitivities.}
  \label{fig:architecture}
\end{figure*}

\subsection{Proposed Solution}

%Overview of the eq. and how to find each component

The required terms in (\ref{eq:opt_summation}) for finding the feature to query includes: the cost of acquiring each feature, the derivative of the prediction function with respect to each input at different input values, and probability of having each value for each feature given the available context. Feature query costs are assumed to be given by the user and known for each time step. For the latter two, while it is possible to model and estimate each term using conventional modeling methods, the solution to evaluate the summation exhaustively may be computationally expensive and impractical in many applications. Here, we introduce a novel method based on autoencoders with binary representation layers that can estimate the whole summation with a single forward and backward propagation in neural networks.

% explain autoencoder with binary layer
The left and the upper right part of the Fig. ~\ref{fig:architecture} show the architecture of the proposed network for the context-aware and one-shot estimation of the distribution of each feature. As depicted, an autoencoder architecture designed to convert each feature in the feature vector ($\bm{x}$) to a binary representation ($\bm{x_{bin}}$). Then, it encodes the features to a more compact representation ($\bm{z}$), and finally reconstructs the original feature vectors  ($\bm{x'}$) by creating a binary decoded vector ($\bm{x'_{bin}}$). Here, in order to have an estimate for the probability of each bit being set, sigmoid non-linearity activation function is used for the binary reconstruction layer ($\bm{x'_{bin}}$). For other activation functions; however, we used the rectified linear unit (ReLu) \cite{nair2010rectified} non-linearity. Additionally, the network optimization cost function is defined as the weighted sum of cross-entropies for binary feature words. \textcolor{r1}{Here, the term word refers to the set of encoded bits that are representing a feature}. The weights are adjusted to offset the importance of the reconstruction error caused by errors in different bits in the word with different significance. It is worth mentioning that the trained autoencoder as explained here, takes an input feature vector where missing features are set to zero, and it is capable of estimating the probability of each bit being set in the binary decode layer ($\bm{x'_{bin}}$).

% \begin{figure*}
% \centering
%   \includegraphics[width=4.5in]{autoencoder}
%   \caption{.}
%   \label{fig:autoencoder}
% \end{figure*}

% explain derivative calculation and the final architecture
In addition to the autoencoder part, in the network of Fig. ~\ref{fig:architecture}, we create a predictor model by stacking a few layers on the top of the encoded representation ($\bm{z}$) and training the encoder as well as the predictor parts of the network in a supervised fashion. Here, in order to measure the sensitivity of the output predictions with respect to different changes in each feature, we suggest using the summation of absolute derivatives of the output layer neurons with respect to each bit of the missing features. The final estimation of the summation in (\ref{eq:opt_summation}) is achieved by an element-wise multiplication of the bit probabilities estimated from the autoencoder's binary reconstruction layer and the sensitivities calculated from the derivative of output layer with respect to each input feature bit.
% final calculation, intuition
Specifically, this paper suggests defining the $RS$ as
\begin{equation}
    RS = \{2^{-l}, \dots, 2^{-2}, 2^{-1}, 1\} \;\;, 
\label{eq:bin_l}
\end{equation}
where $l$ is the total number of bits used in the binary representation of each feature. Using (\ref{eq:opt_summation}) and (\ref{eq:bin_l}), the feature to be acquired is given by
\begin{equation}
    j^{t}_{sel} = \underset{j \in \{1 \dots d\}|k_j^t=0}{argmax} \frac{\sum_{b=1}^{b=l} |\frac{\partial h(\bm{x}^t)}{\partial x_{bin_{j,b}}}|  \: x'_{bin_{j,b}}}{c_j^t} \;\;,
\label{eq:opt_bin}
\end{equation}
where the sensitivity term is defined as
\begin{equation}
    |\frac{\partial h(\bm{x}^t)}{\partial x_{bin_{j,b}}}| = \sum_{i=1}^{i=r} |\frac{\partial y_i}{\partial x_{bin_{j,b}}}| \;\; .
\end{equation}

It is worth mentioning that, in addition to the common neural network hyper-parameters, the only hyper-parameter that is added by the suggested method is the parameter $l$ which is used for controlling the accuracy of the binary representation. Additionally, as we do not make any assumptions on the values used as feature costs and the proposed method is incremental, it can be applied to the scenarios where feature costs are subject to change during the course of operation at test-time. \textcolor{r1}{However, in our experiments, in order to make the comparison of results easier, we evaluate the proposed method on scenarios where feature acquisition costs are constant in time.}

\begin{table*}%
\renewcommand{\arraystretch}{1.3}
\caption{The summary of datasets and experimental settings.}
\label{tab:experiments}
\begin{minipage}{\textwidth}
\begin{center}
\resizebox{\columnwidth}{!}{
\begin{tabular}{cccccccc}
\toprule

\textbf{Dataset} & \textbf{Instances} & \textbf{Features} & \textbf{Classes} & \textbf{Network Architecture} & \textbf{Latent Missing Distribution} \\

\hline
\textbf{MNIST} & 70000 & 784 \footnote{697 features after omitting features with STD of less than $0.0001$ corresponding to margin pixels.} & 10 & Encoder: [697$\times$8, 64, 32] & Beta Distribution\\
\cite{lecun1998mnist} & &  & & Predictor: [16,10]  &  $\alpha = 3.5, \beta = 1.5$\\

\hline
\textbf{Yahoo LTRC} & 34815 & 519 & 5 & Encoder: [519$\times$8, 128, 32] & Beta Distribution \\
\cite{chapelle2011yahoo} & & & & Predictor: [16, 8]  & $\alpha = 1.5, \beta = 1.5$\\

\hline
\textbf{HAPT} & 10929 & 561 & 12 & Encoder: [561$\times$8, 64, 32] & Beta Distribution \\
\cite{reyes2016transition} & & & & Predictor: [16]  & $\alpha = 1.5, \beta = 1.5$\\

\hline
\textbf{Reuters R8} & 7674 & 1000 & 8 & Encoder: [1000$\times$8, 64, 32] & Beta Distribution \\
\cite{lewis1997reuters} & & & & Predictor: [16, 8]  & $\alpha = 3.5, \beta = 1.5$\\

\hline
\textbf{UCI Mushroom} & 8124 & 22\footnote{116 features after one-hot encoding of categorical features.} & 2 & Encoder: [116$\times$8, 16] & Beta Distribution \\
\cite{schlimmer1981mushroom} & &  & & Predictor: [4]  & $\alpha = 5.5, \beta = 1.5$\\

\hline
\textbf{UCI Landsat} & 6435 & 36 & 7 & Encoder: [36$\times$8, 16, 8] & Beta Distribution \\
\cite{blake1998uci} & & & & Predictor: [4]  & $\alpha = 1.5, \beta = 1.5$\\

\hline
\textbf{UCI CTG} & 2126 & 23 & 3 & Encoder: [23$\times$8, 8] & Beta Distribution \\
\cite{ayres2000sisporto} & & & & Predictor: [4]  & $\alpha = 1.5, \beta = 1.5$\\

\hline
\textbf{Synthesized} & 16000 & 64 & 2 & Encoder: [64$\times$8, 16, 10] & Beta Distribution \\
(Section~\ref{sec:synthesized}) & & & & Predictor: [8, 4]  & $\alpha = 1.5, \beta = 1.5$\\

\hline
%\textbf{Diabetes} & 92062 & 44 & 3 & Encoder: [64$\times$8, 32] & Beta Distribution \\
%(Section~\ref{sec:diabetes}) & & & & Predictor: [16]  & $\alpha = 1.5, \beta = 1.5$\\

\textbf{Thyroid} & 279 & 16 & 3 & Encoder: [16$\times$8, 8] & Beta Distribution \\
(Section~\ref{sec:thyroid}) & & & & Predictor: [4]  & $\alpha = 1.5, \beta = 1.5$\\

\bottomrule
\end{tabular}
}
\end{center}
\end{minipage}
\end{table*}%

\subsection{Implementation Details}
\label{sec:implementation}
%initialization, normalizing features, costs, non-linearity, binary code and decode, learning parameters, fine tuning, back-propagation for derivatives , tensorflow\\

Prior to the analysis we have normalized all feature values in the dataset to the range of zero and one. Also, throughout the experiments we used Tensorflow numerical computation library \cite{abadi2016tensorflow} and explored feed-forward neural network architectures. Also, ReLU non-linearity \cite{nair2010rectified} is used for all hidden layers except the binary representation layers. For converting feature values to the suggested binary representation, we implemented the bit by bit recursive conversion in an efficient and parallel manner. Also, for converting back from the binary representation, we implemented the weighted summation of bit values utilizing a fully parallel matrix multiplication, reshape, and addition. In this work, the Adaptive Moment (Adam) optimization algorithm \cite{kingma2014adam} is used to train each network. The Adam hyper-parameters: learning rate ($\alpha$), decay rate for the first moment ($\beta_1$), and decay rate for the second raw moment ($\beta_2$) are set to $0.001$, $0.9$, and $0.999$, respectively.

The process of training the network starts with training the autoencoder part using a weighted cross-entropy loss between the binary representation of the complete feature vectors and the estimated probabilities from the binary reconstruction layer:
\begin{equation}
\begin{aligned}
    f(x,\theta) =  \sum_{j=1}^{d}  \sum_{b=0}^{l} \: 
    2^{-b} \: (\: \: \tilde{x}'_{bin_{j,b}} log(x'_{bin_{j,b}}) \: + \\
    (1-\tilde{x}'_{bin_{j,b}}) log(1-x'_{bin_{j,b}}) \:) \;\; .
\end{aligned}
\end{equation}
In order to train the denoising autoencoder, for each training instance, we sample random values from a latent Beta distribution and use the sampled values as the probability of missing each feature in the training data. After training the autoencoder part, the trained autoencoder network weights are stored, and a few prediction layers are added on top of the encoder part. The reason we store autoencoder weights is that fine-tuning the weights for the prediction task would affect the distribution estimation functionality of the originally trained autoencoder. \textcolor{r1}{In other words, we do fine-tuning for the supervised prediction task, while a copy of the original not fine-tuned autoencoder is used for probabilistic modeling.} Here, to train the predictor network, we use a smaller learning rate ($\alpha=0.0001$) for the pre-trained encoder and a larger learning rate ($\alpha=0.001$) for the new predictor weights. %It is noteworthy to mention that while fine-tuning the encoder part in the supervised step generally increases the prediction performance, the autoencoder weights will not be probability estimators anymore. Accordingly, we store the autoencoder network trained in the first step in order to be used for making probability estimates. 

For the efficient calculation of derivatives we use back-propagation from the values in the output prediction layer to each binary input bit. In this section, we only described the general architecture and training procedures, as we have conducted various experiments on different datasets, the exact network architecture of each case is explained in Section~\ref{sec:Experimental Results}.

%\hlgreen{Algorithm box for training and prediction time\\}

%---------------------------------------------------------
\section{Experimental Results}
\label{sec:Experimental Results}

%----------------------------
\subsection{Datasets and Experiments}
%Architecture, parameters,  (table) (1)\\
The proposed method is evaluated on seven different real-world datasets including human activity recognition (HAPT) \cite{reyes2016transition}, hand-written character recognition (MNIST) \cite{lecun1998mnist}, document classification (Reuters R8) \cite{lewis1997reuters}, and web ranking (Yahoo LTRC) \cite{chapelle2011yahoo} as well as three other classification datasets. Apart from these, we have evaluated the method on a synthesized dataset which is explained in Section~\ref{sec:synthesized} and a dataset in health domain explained in Section~\ref{sec:thyroid}. Table~\ref{tab:experiments} presents a summery of the conducted experiments. The table also includes the network architecture and the missing value distribution used during the training phase. In each case, the architecture column contains encoder layer sizes for the binary layers, encoder layers, and predictor layers. In this table, the decoder layers are not shown and are equal to the encoder layer sizes in reverse order. We used an 8-bit binary representation throughout the experiments. Regarding the feature size of each dataset, we report both the nominal feature count and the number of features we have used in our experiments. Specifically, for the MNIST dataset, each pixel is considered as a feature and we removed features corresponding to pixels near the margin that are almost always zero with the standard deviation of less than $0.0001$ across all samples. Also, regarding the Mushroom dataset, one-hot encoding of $22$ categorical features resulted in $116$ features to be used as input. An important point to consider for these features is that, during sensitivity measurement and acquisition, we should consider acquisition of all one-hot features corresponding to a categorical feature as a single feature to acquire.

Regarding the feature acquisition costs, for the LTRC dataset, as suggested by \cite{xu2014classifier}, we used real cost values in the range of $1$ to $150$ based on the time required to extract each feature. In order to introduce feature costs to the MNIST dataset, we have followed the method suggested by \cite{trapeznikov2013supervised}. Using this method, we create feature vectors by concatenating MNIST images at different resolutions including $28\times28$, $14 \times 14$, $7 \times 7$, and $4 \times 4$ with feature acquisition costs of $4$, $3$, $2$, and $1$ for acquiring feature at each resolution, respectively. For the Landsat dataset, each sample consists of features from four different frequency bands. Here, we considered features of the same frequency band to have the same acquisition cost from $1$ to $4$, equal to the frequency band number. At last, for the CTG dataset, we assumed that features that are measuring event counts to have cost value of $1$, features measuring statistical information to have the cost value of $2$, and histogram features to have the cost of $3$. For the synthesized and diabetes datasets, a complete explanation experiments is presented in Section~\ref{sec:synthesized} and Section~\ref{sec:thyroid}, respectively. Lastly, for the other three datasets, we assumed feature costs to be equal for all features.

Based on the aforementioned experimental setup, in the following parts of this section, we evaluate the proposed method for feature acquisition considering costs at test-time (FACT) on each dataset.

%----------------------------
\subsection{Performance Evaluation}

\begin{table*}%
\renewcommand{\arraystretch}{1.3}
\caption{Results of evaluating the proposed method on different datasets.}
\label{tab:eval}
\begin{minipage}{\textwidth}
\begin{center}
\resizebox{\columnwidth}{!}{
\begin{tabular}{ccccccccccc}
\toprule

\textbf{Dataset} & \multicolumn{5}{c}{\textbf{FACT Accuracy} (\%)} & \textbf{RFC Accuracy} & \textbf{Denoising} & \textbf{Rand} & \textbf{\textcolor{r1}{DPFQ}\footnotemark[3]} & \textbf{FACT}\\
  & \multicolumn{5}{c}{\% of total cost used\footnotemark[1]} & (\%) & (\%) & (AUACC\footnotemark[2]) & (AUACC) & (AUACC)\\
 & $0\%$ & $25\%$ & $50\%$ & $75\%$ & $100\%$ & $100\%$ total cost\footnotemark[1] & & & \\
\hline

\textbf{MNIST} & $10.63$ & $94.95$ & $96.19$ & $96.17$ & $96.17$ & $97.07$ & $60.08$ & $0.63$ & $0.71$ & $0.82$ \\

\textbf{Yahoo LTRC} & $20.70$ & $47.16$ & $49.41$ & $50.48$ & $50.39$ & $50.51$ & $94.03$ & $0.47$ & $0.47$ & $0.48$ \\

\textbf{HAPT} & $16.60$ & $87.57$ & $90.22$ & $90.71$ & $90.77$ & $91.75$ & $90.09$ & $0.65$ & $0.69$ & $0.76$ \\

\textbf{Reuters R8} & $50.74$ & $94.78$ & $95.56$ & $94.78$ & $94.70$ & $94.61$ & $14.92$ & $0.87$ & $0.89$ & $0.93$ \\

\textbf{UCI Mushroom} & $50.41$ & $99.26$ & $99.83$ & $99.91$ & $99.93$ & $99.99$ & $57.69$ & $0.61$ & $0.85$ & $0.85$ \\

\textbf{UCI Landsat} & $27.66$ & $80.10$ & $83.73$ & $87.35$ & $88.49$ & $91.08$ & $93.88$ & $0.65$ & $0.68$ & $0.72$ \\

\textbf{UCI CTG} & $74.52$ & $84.90$ & $88.68$ & $88.99$ & $90.88$ & $89.62$ & $77.48$ & $0.83$ & $0.84$ & $0.86$ \\

\textbf{Synthesized} & $50.43$ & $96.72$ & $98.19$ & $98.25$ & $98.33$ & $99.41$ & $88.78$ & $0.64$ & $0.73$ & $0.78$ \\

%\textbf{Diabetes} & $39.51$ & $72.07$ & $75.60$ & $75.78$ & %$75.78$ & $70.85$ & $70.94$ & $0.51$ & $0.54$ & \bm{$0.63$} \\

\textbf{Thyroid} & $26.83$ & $65.85$ & $70.73$ & $75.61$ & $78.05$ & $80.85$ & $44.07$ & $0.59$ & $0.59$ & $0.71$ \\

\bottomrule
\end{tabular}
}
\end{center}
\begin{flushleft}
\footnotemark[1] Total cost defined as the total cost of acquiring all features.\\
\footnotemark[2] AUACC is defined as the area under the accuracy and the feature acquisition cost curve.\\
\footnotemark[3] Cost-aware version of the method suggested in \cite{kachuee2017context}. %which is basically FACT without the binary encoding.
\end{flushleft}
\end{minipage}
\end{table*}%

%number of queries Vs accuracy curves (table) (2), final prediction performance on complete feature vectors and comparison with simple ML methods (table) (2)\\

% how experiments designed
 We split each dataset into three parts: $15\%$ for test, $15\%$ for validation, and the rest for training. During the training phase, we use the  validation and train sets to train the networks as explained in Section~\ref{sec:implementation} and following the setups introduced in Table~\ref{tab:experiments}. \textcolor{r1}{It is worth noting that the proposed method does not necessary require all training features to be known. In fact, as explained in Section~\ref{sec:implementation}, we use a latent Beta distribution to simulate the existence of unknown features.} After the training phase, we use the test set to simulate the case of feature acquisition using the proposed method by assuming all the features to be initially unknown and using the feature query criterion of (\ref{eq:opt_bin}) to ask for features incrementally. In our experiments (except experiments in Section~\ref{sec:synthesized}), we continue the feature query until querying all the features and report the accuracy as well as the cost at each point during feature acquisition. In real applications; however, the incremental acquisition should be stopped after reaching a certain criterion such as a minimum confidence of predictions.

% summary of table columns
Table~\ref{tab:eval} presents the results of the proposed method on each dataset. The table contains the test accuracy of each dataset while asking \textcolor{r1}{for different percentage of the total cost from the original feature set. Here, total cost is defined as the cost of acquiring all features}. As a baseline, we have reported the test performance of using a random forest classifier (RFC) on the complete feature set \textcolor{r1}{(i.e., acquiring all features and spending the maximum cost)}. The table also reports the denoising percentage of the trained denoising autoencoder calculated as
\begin{equation}
100 \:\times\: \frac{||\bm{x}-\bm{\tilde{x}}|| - ||\bm{x'}-\bm{\tilde{x}}||}{||\bm{x}-\bm{\tilde{x}}||} \;\; .
\end{equation}
Additionally, in this table, \textcolor{r1}{the area under the accuracy-cost curves (AUACC) of the proposed feature query method as well as the AUACC of randomly asking for the unknown features are presented. We have also included AUACC results of a cost-aware version of the method suggested in \cite{kachuee2017context} in which sensitivities are normalized by feature costs (see the DPFQ column). It is worth mentioning that AUACC values are calculated as the normalized area under the accuracy versus the acquisition cost curve from a cost of zero to the cost at which the accuracy converges to the maximum accuracy.}

% analysis of table values
According to the results presented in Table~\ref{tab:eval}, the proposed method can be used to effectively reduce the cost of features to be acquired at test-time for accurate predictions. Also, comparing the baseline accuracy of using the complete feature set, the results show that our method trains predictor models that can make viable predictions using only a subset of the features without sacrificing prediction performance. Regarding the denoising percentage, in most of the cases, \textcolor{r1}{the achieved denoising percentage is significant and confirming that a denoising autoencoder is capable of encoding features to reduce the feature representation length} which results in a new representation that is more robust to the presence of missing values. Regarding the area under the curve values, in all cases, the AUC values of using FACT is considerably higher than its random selection counterpart. \textcolor{r1}{It is also noteworthy to mention that for a few datasets (i.e., Reuters R8, UCI Landsat, HAPT, and UCI CTG) there is a considerable class imbalance that is affecting baseline accuracies.}

\begin{figure*}[t!]
    \centering
    \begin{subfigure}[b]{0.80\textwidth}
        \centering
        \includegraphics[width=0.97\linewidth]{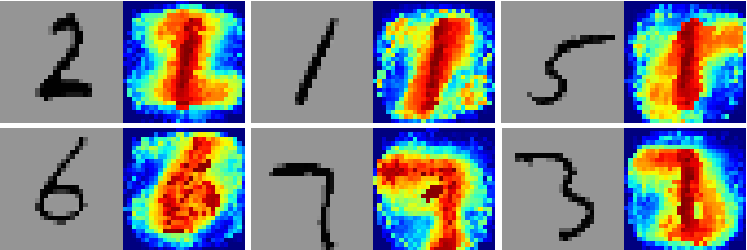}
        \caption{}
        \label{fig:mnist_vis_dynamic}
    \end{subfigure}%
    ~ 
    \begin{subfigure}[b]{0.16\textwidth}
        \centering
        \includegraphics[width=\linewidth]{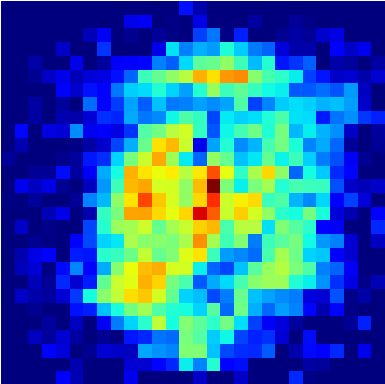}
        \caption{}
        \label{fig:mnist_vis_static}
    \end{subfigure}
    
    \vspace{0.25in}
    
    \begin{subfigure}[b]{0.95\textwidth}
        \centering
        \includegraphics[width=\linewidth]{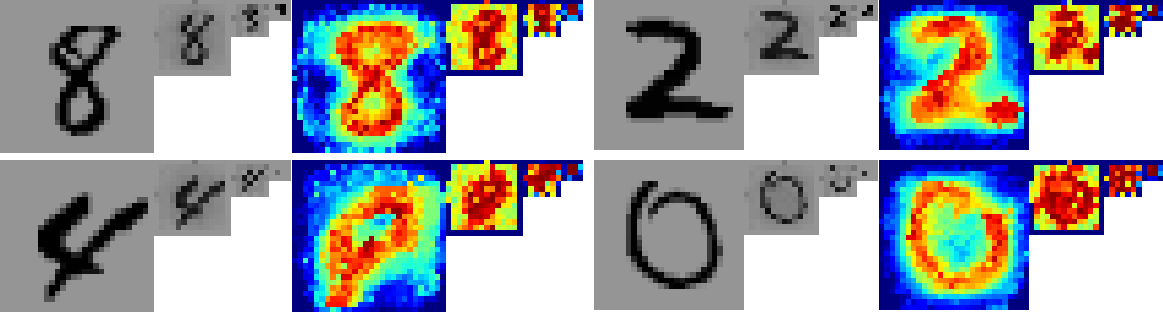}
        \caption{}
        \label{fig:mnist_vis_multires}
    \end{subfigure}
    \caption{Visualization of: (a) using the proposed approach on the MNIST dataset with equal feature costs, (b) using static feature acquisition using mutual information between pixels and targets on MNIST dataset, and (c) using the proposed approach on the multi-resolution MNIST dataset with different feature costs at each resolution. Pixels with more importance/priority to be queried are indicated by warmer colors.}
    \label{fig:mnist_vis}
\end{figure*}

%MNIST visualization \\
To further illustrate the performance of the proposed approach, we used MNIST dataset to visualize the effect of cost and context on the selection of features to be queried. Here, features are pixel values at different locations across each image, and the context is the available pixel values at each time step. Fig.~\ref{fig:mnist_vis} shows the effect of context and cost on the order of features that are queried by the proposed algorithm as well as a static order which is measured based on mutual information between pixels and target classes. Here, we present results for the original MNIST dataset with single resolution images and equal feature costs (see Fig.~\ref{fig:mnist_vis_dynamic}) as well as the introduced multi-resolution setup with different feature costs for each resolution (see Fig.~\ref{fig:mnist_vis_multires}). In this figure, pixels with higher importance to be queried are indicated by warmer colors and less important pixels are indicated by colder colors. As it is evident from this figure, the proposed context-aware method, based on the available pixels, acquires features with different orders and is scanning for digit edges or discriminative areas. On the other hand, the static feature acquisition method, only asks for central pixels of each image in a fixed order (see Fig.~\ref{fig:mnist_vis_static}). In addition, regarding the multi-resolution case, as it can be seen from the figure, the informative pixels from lower resolutions that incur lower costs are preferred to more costly higher resolution pixels. For instance, in the lower left corner image of Fig.~\ref{fig:mnist_vis_multires}, the parts that are first acquired from the high-resolution pixels are the parts that create difference between the digits of 4 and 9 which are not clear enough in lower resolutions.

%Normalized accuracy plot of using NBITS from 2 to 32 (one multicol plot) (3)\\
% NA.

%----------------------------
\subsection{Comparison with Other Work}

\begin{figure}[t]
\centering
  \includegraphics[width=1.03\linewidth]{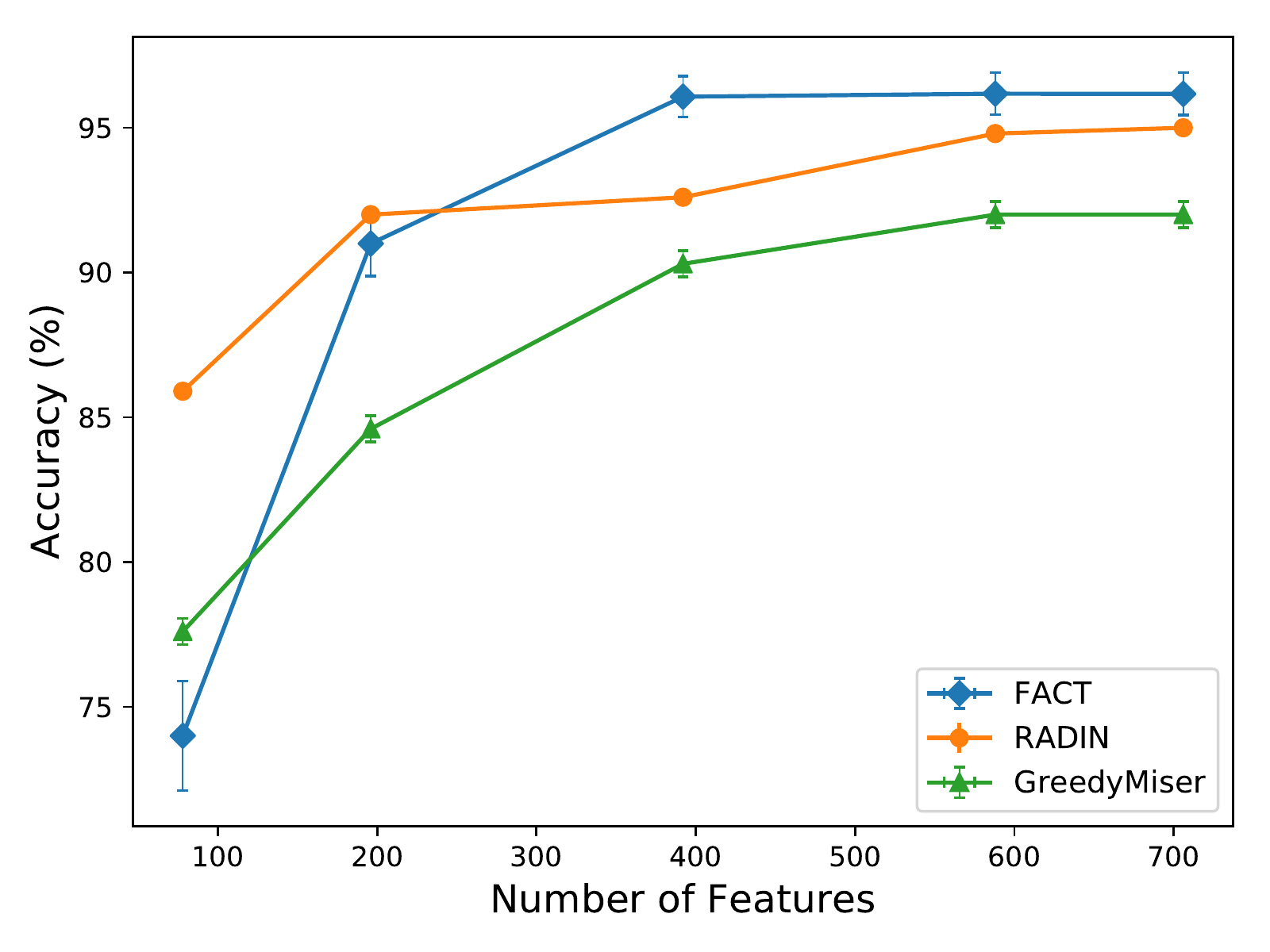}
  \caption{Comparison of the proposed method (FACT) with RADIN \cite{contardo2016recurrent} and GreedyMiser \cite{xu2012greedy} methods on the MNIST dataset. %The X-axis represents the number of acquired features and Y-axis represents the accuracy achieved using each method.
  }
  \label{fig:comp_mnist}
\end{figure}

Fig.~\ref{fig:comp_mnist} presents comparison of the proposed feature acquisition method with a feature acquisition method based on recurrent neural networks (RADIN) \cite{contardo2016recurrent} and a tree-based feature acquisition method (GreedyMiser) \cite{xu2012greedy}. In this comparison, we have used MNIST dataset to evaluate the performance of each method based on their accuracy using a different number of features. As it can be inferred from the figure, in the case of acquiring $10\%$ of features, where the number of features to be queried is significantly less than the total number of features, the achieved accuracy using FACT is lower than other methods. Nevertheless, the rate of increase in the accuracy with respect to the number of queried features for the presented method is significantly higher than other papers which makes it superior in other cases. \textcolor{black}{In this plot and similar cost versus accuracy plots in this section, we provide 95\% confidence intervals presented as error-bars measured by running each experiment multiple times using different random initializations.}

% comparison with RADIN and GreedyMiser on MNIST and R8
% \begin{figure*}[t!]
%     \centering
%     \begin{subfigure}[c]{0.48\textwidth}
%         \centering
%         \includegraphics[width=\linewidth]{comp_mnist}
%         \caption{}
%         \label{fig:comp_radin_mnist}
%     \end{subfigure}%
%     ~ 
%     \begin{subfigure}[c]{0.48\textwidth}
%         \centering
%         \includegraphics[width=\linewidth]{comp_r8}
%         \caption{}
%         \label{fig:comp_radin_r8}
%     \end{subfigure}
%     \caption{Comparison between the proposed method (OPFA) with RADIN \cite{contardo2016recurrent} and GreedyMiser \cite{xu2012greedy} methods on (a) MNIST and (b) Reuters R8 dataset. The x-axis represents the number of acquired features and y-axis represents the accuracy achieved using each method.}
%     \label{fig:comp_radin}
% \end{figure*}

\begin{figure}[t]
\centering
  \includegraphics[width=\linewidth]{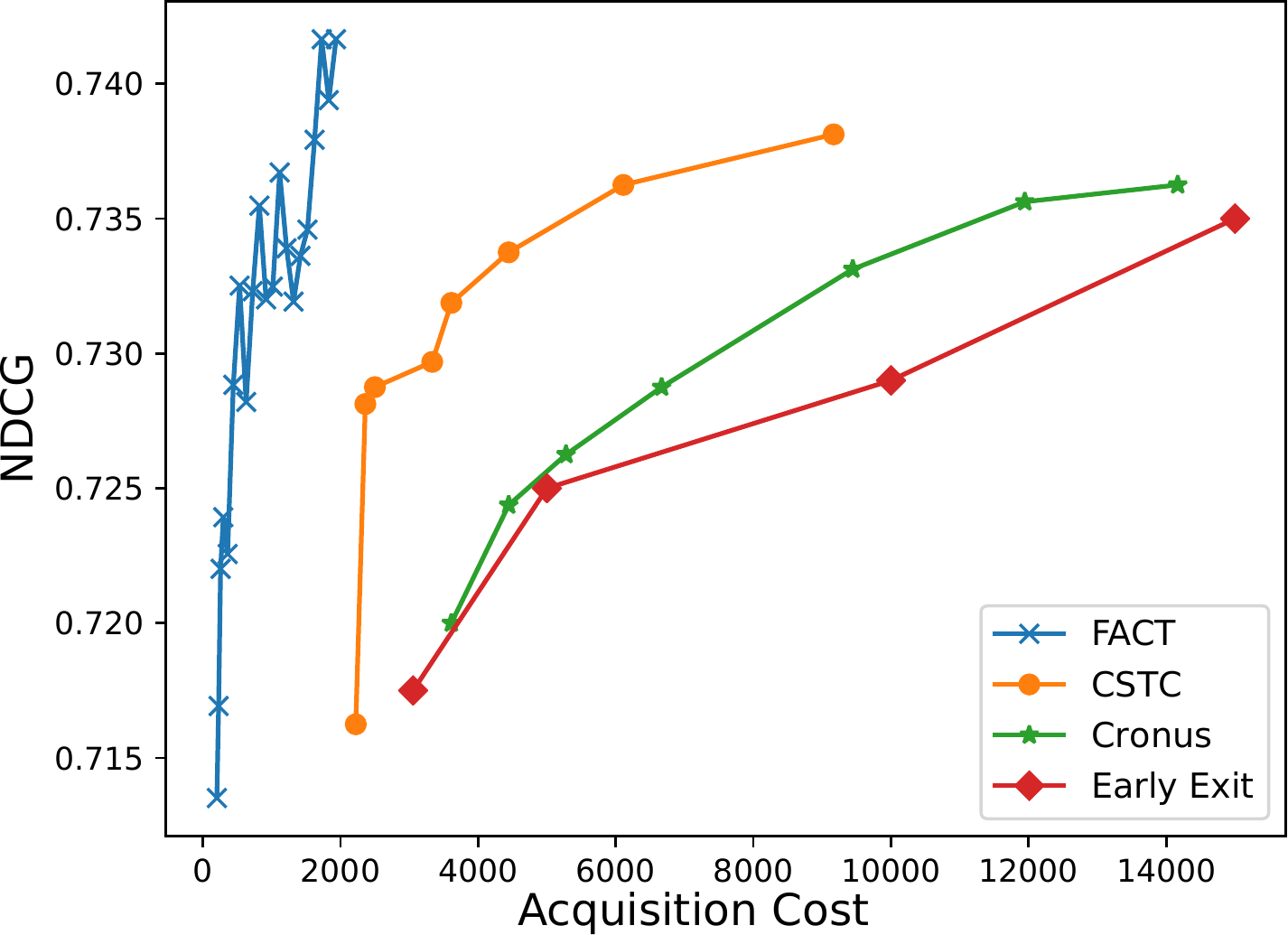}
  \caption{Comparison of the proposed method (FACT) with CSTC \cite{xu2014classifier}, Cronus \cite{chen2012classifier}, and Early Exit \cite{cambazoglu2010early} methods on the LTRC dataset.}
  \label{fig:comp_ltrc}
\end{figure}

%NDGC on LTRC and CSTC, Cronus, EarlyExit comparison (plot) (1) \\
Fig.~\ref{fig:comp_ltrc} presents a comparison between the feature acquisition and cost curve of the proposed method with three other approaches that use classifier cascades or trees in the literature including CSTC \cite{xu2014classifier}, Cronus \cite{chen2012classifier}, and Early Exit \cite{cambazoglu2010early}. Here, we used the LTRC dataset with the feature costs as suggested by \cite{xu2014classifier} to plot the feature acquisition cost versus the corresponding normalized discounted cumulative gain (NDCG) \cite{jarvelin2002cumulated} performance measure. NDCG is a well-known measure of ranking quality that is used to measure the effectiveness and the relevance of ranking results in search engines. As it can be seen from this figure, FACT is significantly more powerful and more efficient compared to others.

%\subsubsection{Comparison with the Sensitivity-Based Method}

\begin{figure}[t]
\centering
  \includegraphics[width=\linewidth]{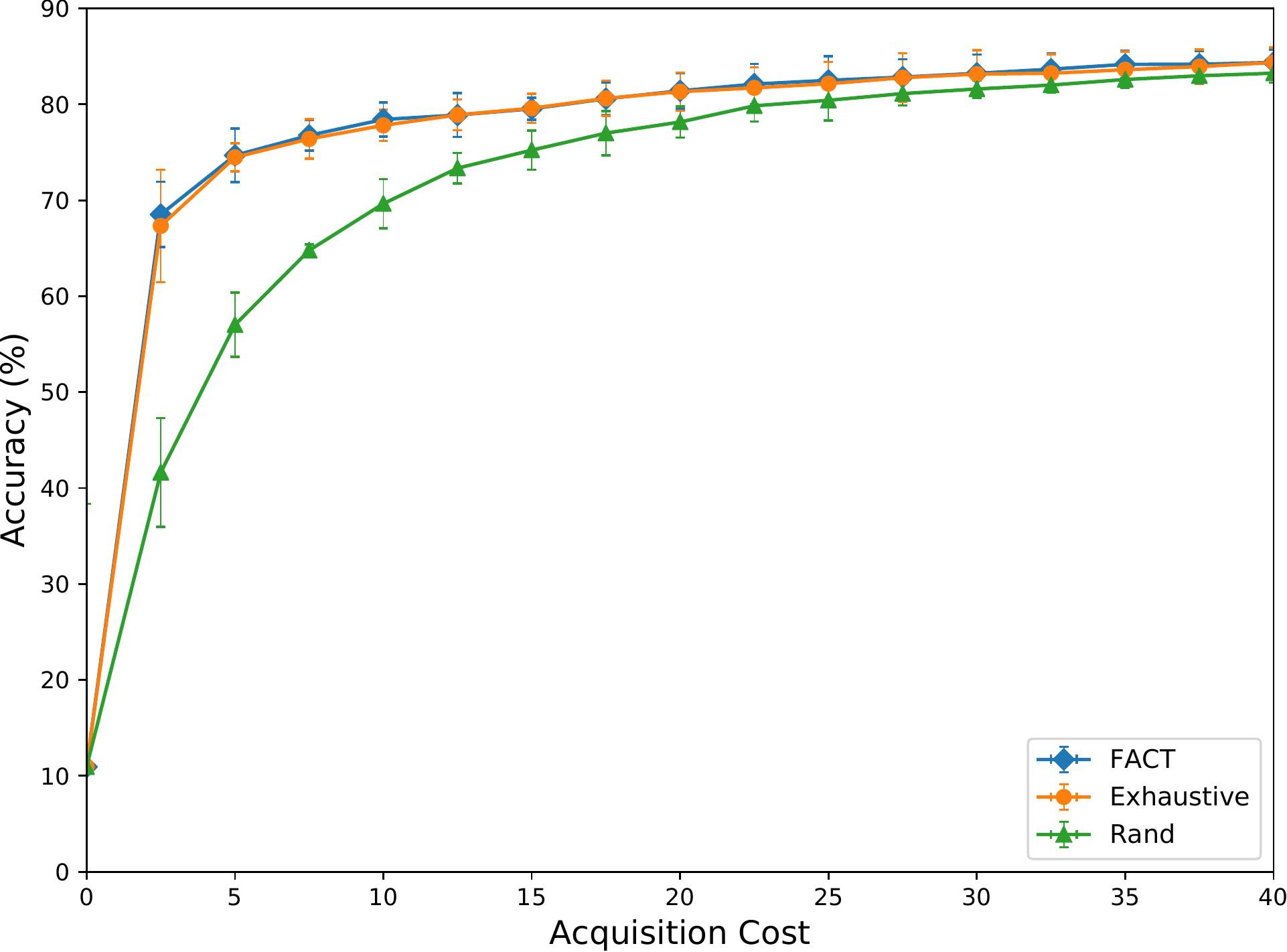}
  \caption{Comparison of the proposed method (FACT) with exhaustive sensitivity-based (Exhaustive) \cite{early2016test} and random selection (Rand) methods on the Landsat dataset. This figure shows that the proposed method is able to approximate the ground-truth sensitivity values accurately and efficiently.}
  \label{fig:comp_exhaustive_satimage}
\end{figure}

In order to evaluate the performance of the proposed method for the estimation of sensitivity values, we have implemented an exhaustive feature query method as suggested by \cite{early2016test} using sensitivity as the utility function. To make a fair comparison, we have used the trained predictor network, and exhaustively measured the effect of changing each input on the prediction probabilities. Here, in order to estimate the probability of each change, we have used a 5-bin histogram for each feature.
Fig.~\ref{fig:comp_exhaustive_satimage} presents a comparison between the accuracy achieved using FACT and the exhaustive sensitivity-based method on the Landsat dataset. As a baseline, we have also included the curve corresponding to randomly selecting and acquiring features. As it is evident from the figure, the proposed method is almost equivalent to the exhaustive method in terms of the accuracy achieved at each total acquisition cost. This is promising considering the fact that the proposed approach tries to approximate the exhaustive sensitivity measurement in an efficient and scalable manner. In other words, compared to the proposed method, the exhaustive method is significantly slower and less efficient at test-time. Specifically, the average processing time of the exhaustive method was about $2400~ms$ for each sample, while the corresponding processing time for FACT was about $20~ms$ (i.e., about $120$ times faster). In other test cases with more features, comparing with the exhaustive method was not possible due to the exponentially increasing computational load of evaluating the exhaustive method. 

It worth mentioning that this computational advantage comes from the fact that, for each incremental feature query, we approximate the summation of (\ref{eq:opt_bin}) for all unknown features using one forward and one backward network computation. However, the exhaustive sensitivity measurement computes the summation for each feature and over the range of all possible values, separately. In other words, the proposed method scales linearly with the growth in the number of unknown features, while the exhaustive method scales in polynomial time.

% NA.
%Comparison with newer papers: reinforcement learning paper on energy estimate and student life (2) \\
%Computational load comparison with pure sensitivity (3)\\

\subsection{Evaluation using Synthesized Data}

\label{sec:synthesized}

\begin{figure}[p]
    \centering
    \begin{subfigure}[b]{\columnwidth}
        \centering
        \includegraphics[width=\columnwidth]{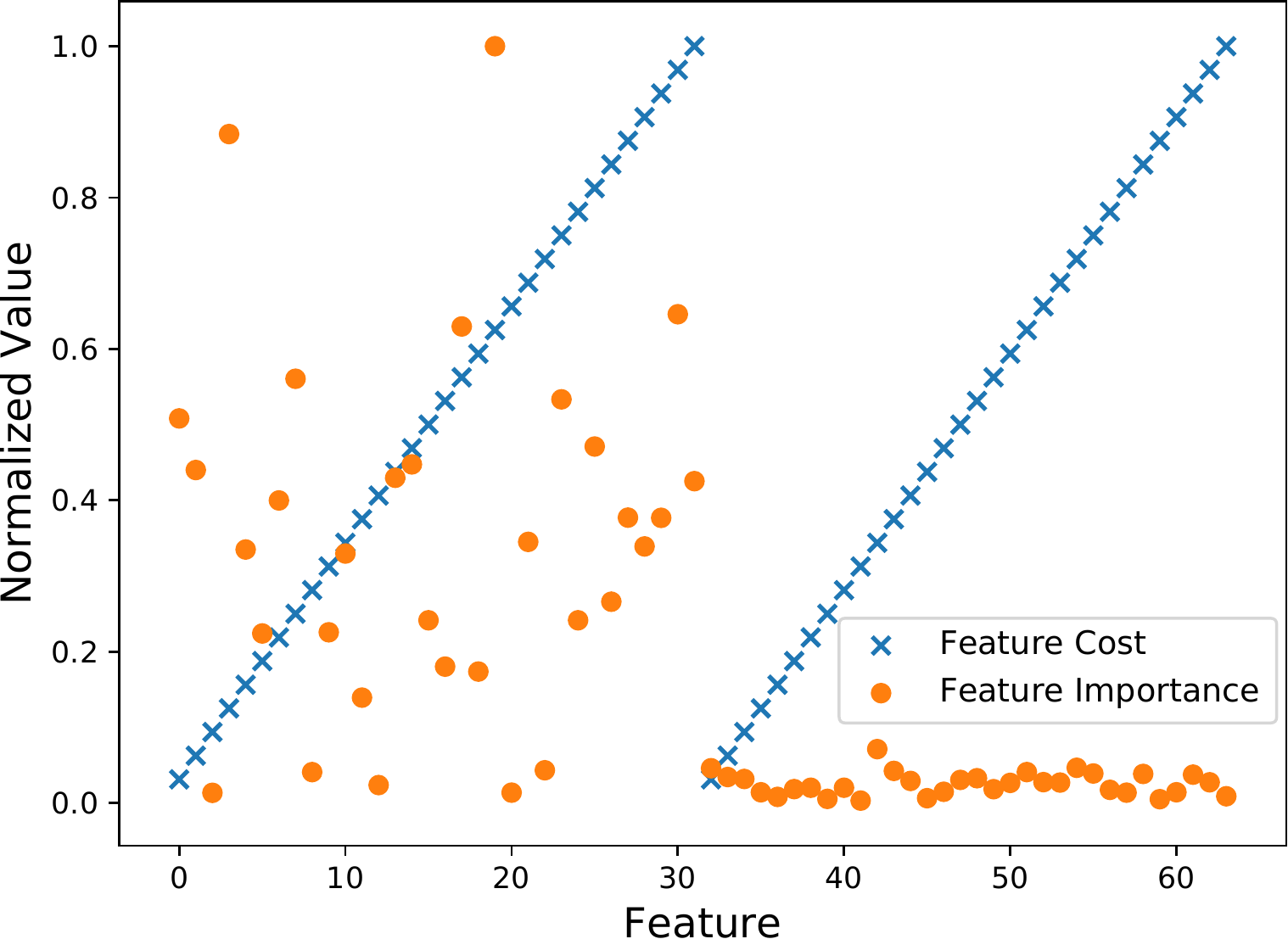}
        \caption{}
        \label{fig:syn_eval_fecost}
    \end{subfigure}%
    
    \begin{subfigure}[b]{\columnwidth}
        \centering
        \includegraphics[width=\columnwidth]{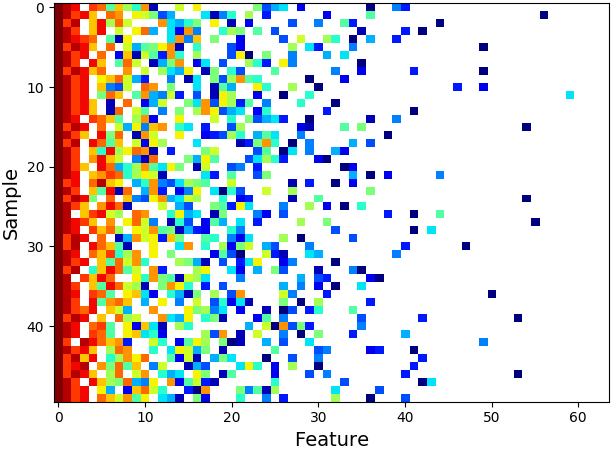}
        \caption{}
        \label{fig:syn_eval_order}
    \end{subfigure}
    
    \begin{subfigure}[b]{\columnwidth}
        \centering
        \includegraphics[width=1.03\columnwidth]{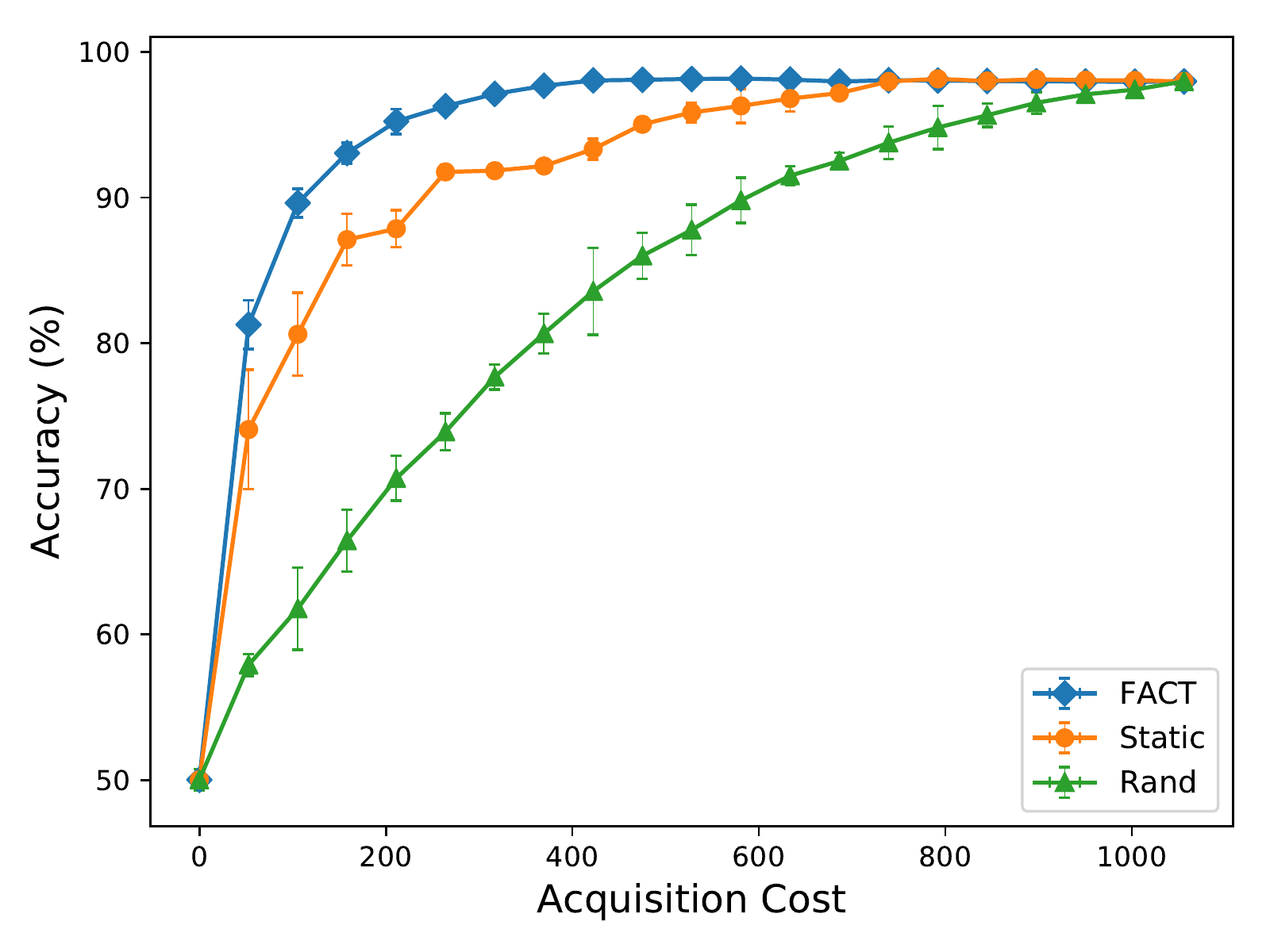}
        \caption{}
        \label{fig:syn_eval_curve}
    \end{subfigure}
    \caption{Evaluation of the proposed method on synthesized data. (a) Cost and static importance of each feature. (b) The feature acquisition order for 50 different test samples (warmer colors mean more priority). (c) Accuracy versus acquisition cost curves for the proposed method (FACT), acquisition using static order, and random selection.}
    \label{fig:syn_eval}
\end{figure}

In order to get more insight about the performance of the proposed method, we have used a synthesized dataset to evaluate the suggested approach. The synthesized dataset is generated as follows: first, we have randomly sampled $16$ cluster centers from a $32$-dimensional space. Then, $1000$ points sampled around each cluster center from a normal distribution with the mean of zero and variance of $0.25$. Afterwards, we have randomly assigned each cluster to a class from a set of two different classes. To each feature vector created so far containing $32$ features, we have appended another $32$ features with random values from a normal distribution. These are features without any predictive value. Accordingly, the resulting feature vectors are of size $64$. Finally, we made the dataset cost-sensitive by defining feature costs for the first and second $32$ features to be a monotonically increasing function from $1$ to $32$. See Fig.~\ref{fig:syn_eval_fecost} for a visualization of feature costs and the static importance computed using the mutual information between features and labels. %Using this distribution of features and costs, in the feature acquisition process, we have unknown features with the same cost having different level of information.

Fig.~\ref{fig:syn_eval_order} demonstrates the order in which each feature is acquired using the proposed method. In this figure, each row corresponds to a test sample (here only 50 samples are visualized) and each column represents a feature. The features that are acquired earlier are indicated with warmer colors. Here, we have continued the feature acquisition until we reach $95\%$ of the maximum achievable accuracy. As it can be seen, the second half of features which are not informative are mainly skipped by the proposed method. Specifically, only about $5.2\%$ of features from the second half are selected by FACT, which means that most of the uninformative features are not acquired by the algorithm. On the other hand, based on the cost and value of each feature from the first half, the proposed method acquired features that are more informative and have lower cost values. Apart from this, Fig.~\ref{fig:syn_eval_curve} presents the accuracy versus total feature acquisition cost on the test set. As it can be seen from the curve, FACT converges to the maximum accuracy much faster than the static and random acquisition methods. It is mainly due to the fact that the proposed method highly prefers informative features with low cost while other methods disregard this information.

\subsection{\textcolor{r1}{Evaluation using Real-World Health Data}}

\label{sec:thyroid}

In order to evaluate the performance of the proposed method on a dataset in health domain where feature acquisition costs are inherently important, we have used thyroid classification dataset \cite{Dua2017}\footnote{Available at: http://archive.ics.uci.edu/ml/datasets/thyroid+disease}. Here, we have features from different categories including demographics, questionnaire, examination, and lab results. Furthermore, this dataset provides the acquisition costs corresponding to each feature which ranges from $1.0$ for features such as age to $22.78$ for certain blood tests.

\begin{figure}[t]
    \centering
    
    \begin{subfigure}[b]{\columnwidth}
        \centering
        \includegraphics[width=\columnwidth]{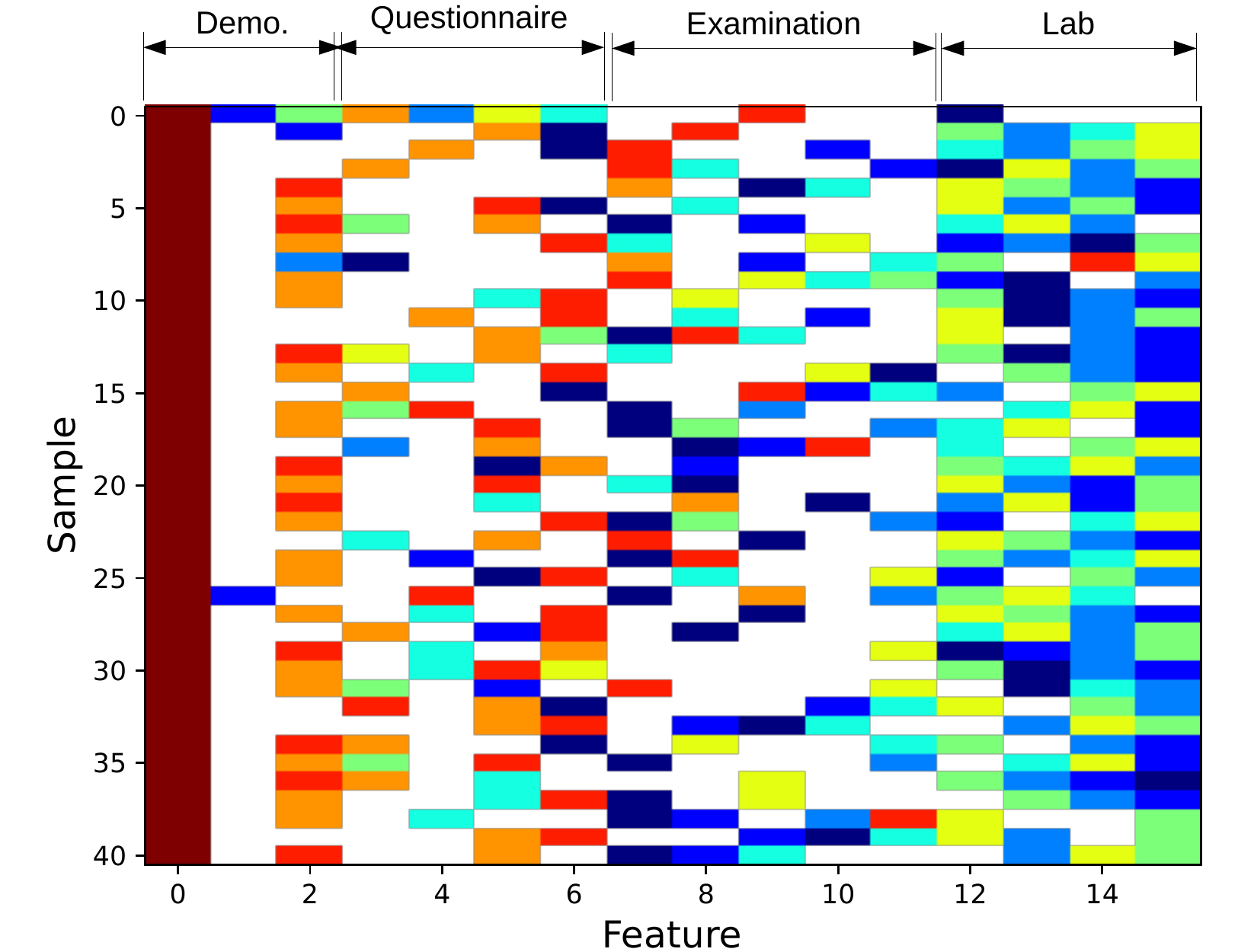}
        \caption{}
        \label{fig:thyroid_eval_order}
    \end{subfigure}
    
    \begin{subfigure}[b]{\columnwidth}
        \centering
        \includegraphics[width=\columnwidth]{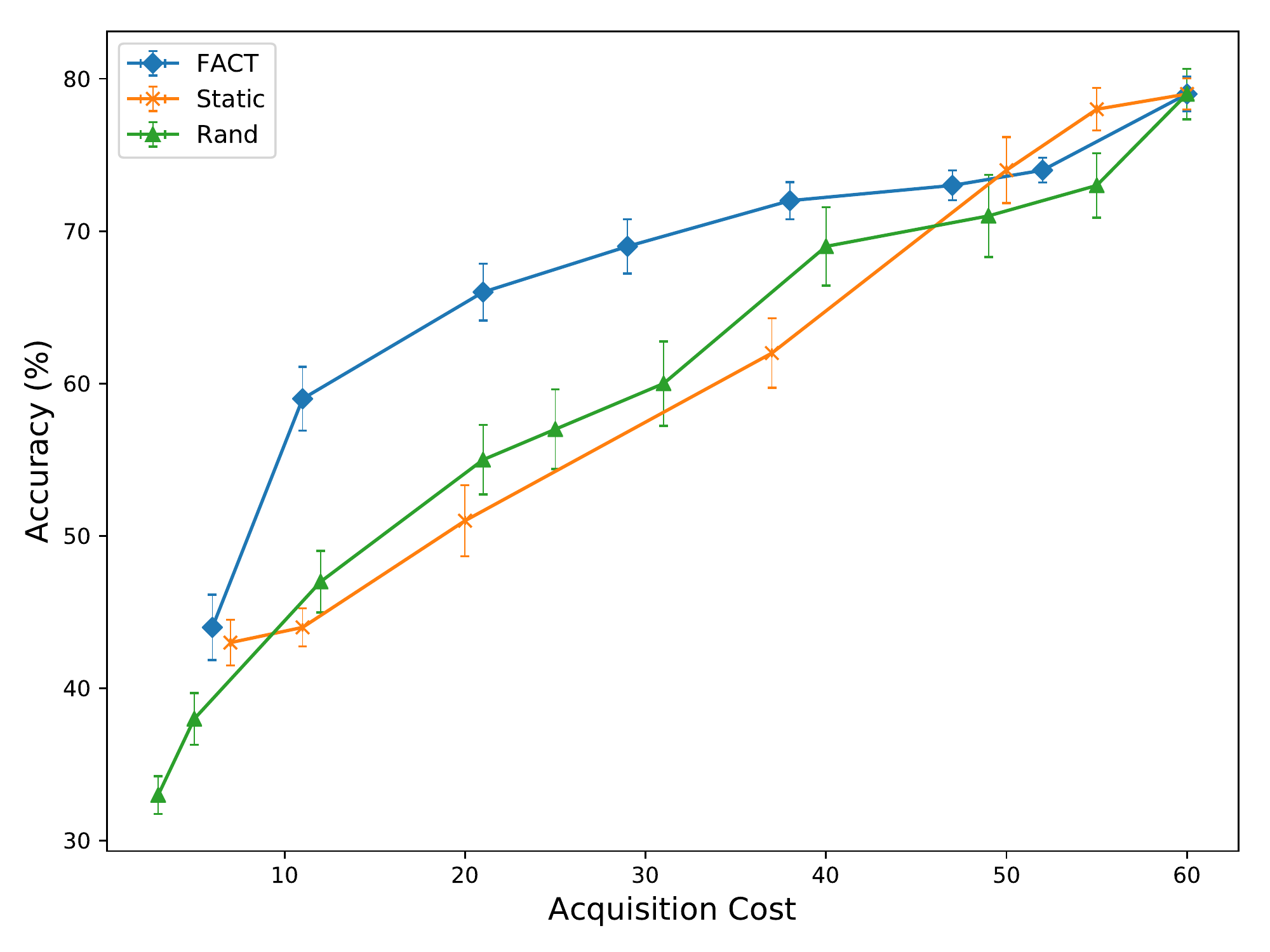}
        \caption{}
        \label{fig:thyroid_eval_curve}
    \end{subfigure}
    \caption{Evaluation of the proposed method on thyroid disease classification task. (a) The feature acquisition order for 50 different test samples (warmer colors mean more priority). (b) Accuracy versus acquisition cost curves for the proposed method (FACT), acquisition using static order, and random selection.}
    \label{fig:thyroid_eval}
\end{figure}

Figure~\ref{fig:thyroid_eval_order} presents a visualization of orders which each feature is acquired for 40 randomly selected test samples. As it can be observed from this visualization, FACT gives more priority to low cost features and costly but informative features are acquired a with lower priority. 
Apart from this, Figure~\ref{fig:thyroid_eval_curve} presents the accuracy versus acquisition cost curve for FACT, static, and random methods. As it can be seen from this figure, FACT outperforms other baseline approaches.

\section{\textcolor{r1}{Discussion}}

\begin{figure}[t]
\centering
  \includegraphics[width=\linewidth]{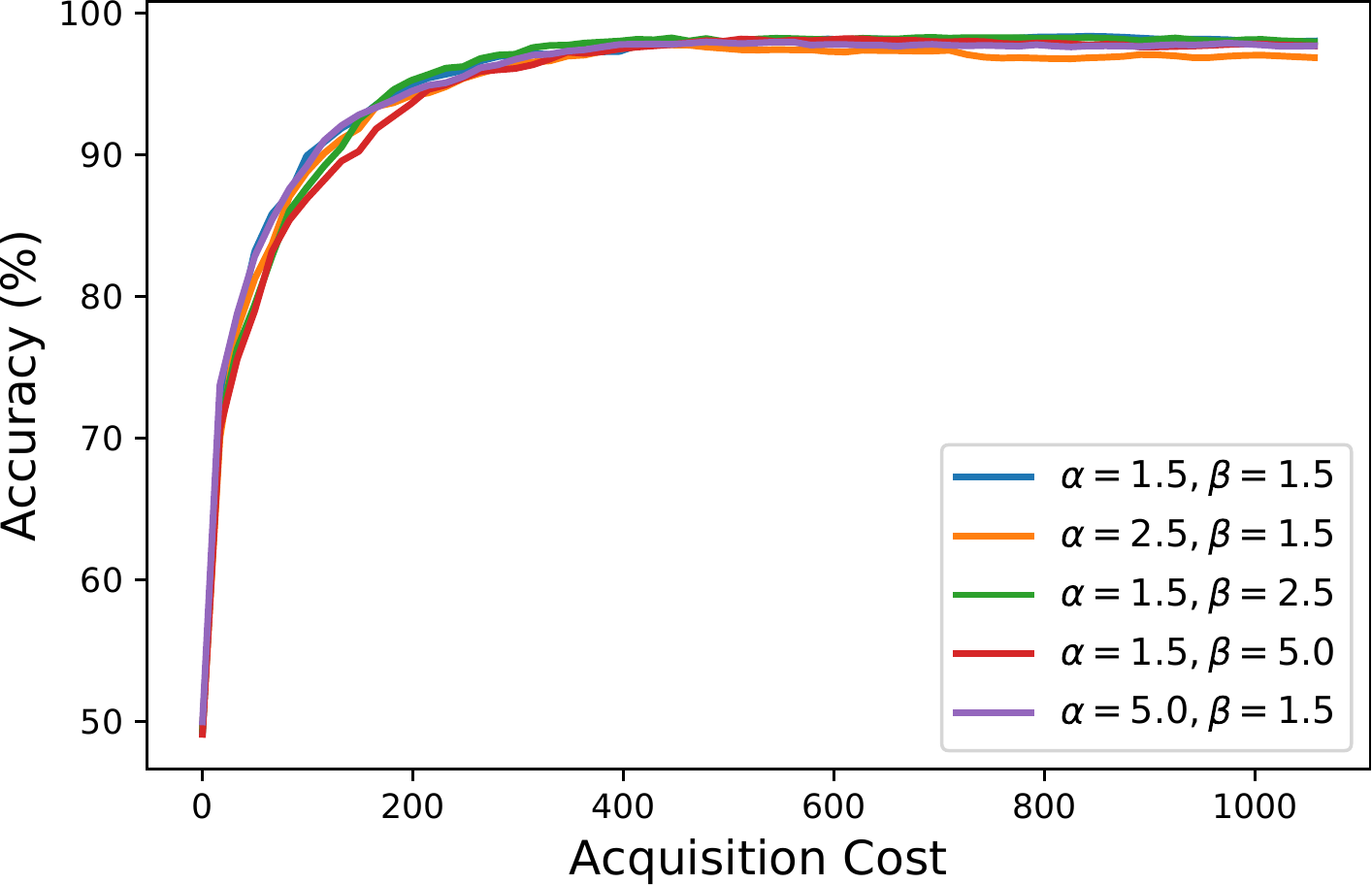}
  \caption{Influence of different beta distribution parameters on the accuracy versus acquisition cost curve for the synthesized dataset.}
  \label{fig:comp_beta}
\end{figure}

There are many methods such as mutual information, information gain etc. that are traditionally used in literature to measure the value of each feature \cite{liu1995chi2}. However, these methods are usually limited to considering only linear relationships or considering a single feature rather than joint distribution of features. For instance, given evidence about a subset of features, the correlations between the rest of features maybe affected that traditional approaches are usually incapable of capturing. In this paper, we suggest inferring the dynamics between features and classes by sensitivity analysis of trained predictors. This approach employs the hidden information captured inside a black-box network to measure the value of acquiring each feature given the available context. \textcolor{black}{In this paper, we use \eqref{eq:opt_sense} as a measure of feature informativeness per unit of the cost to make feature acquisition decisions. However, an alternative approach, which may result in better accuracies at a certain context, would be defining an objective function that balances the cost versus performance trade-off using a hyper-parameter.}

Furthermore, this paper suggests an encoding and decoding approach to create a range of changes so that the final summation of sensitivities would be a better approximation of the total sensitivity with respect to each feature. Aside from binary quantization, we have explored different methods such as variable length and constant length quantizations; however, while they usually work reasonably well, we decided to use binary encoding as it is more efficient to implement and our readers are more familiar with.

In this paper, a beta corruption function is used to introduce missing features and to train the denoising autoencoder. Based on our experiments, as long as it is chosen reasonably, it does not have any direct influence on the performance of the predictor or the feature acquisition functionality. Specifically, we measured the influence of changing beta parameters from $1$ to $5$ and the area under the accuracy cost curve changes were less than $1\%$ (see Fig.~\ref{fig:comp_beta} for an example). In this paper, we suggest beta distribution parameters of $\alpha{=}1.5, \beta{=}1.5$ for most datasets, and parameters of $\alpha{=}5.5, \beta{=}1.5$ sparse datasets such as mushroom. It is also worth mentioning that the corruption function is applied to all features independently. Therefore, it does not introduce any bias toward certain features.

%---------------------------------------------------------
\section{Conclusion}
\label{sec:Conclusion}
%Summary of the suggested method \\
%Its features: context-aware, cost-aware, scalable \\
%Results summary \\

In this paper, we introduced a novel method for cost- and context-aware feature acquisition at test-time. The proposed method based on denoising autoencoders with binary representation layers efficiently estimates context-dependent feature distributions and measures the sensitivity of the output with respect to each unknown feature. Furthermore, we evaluated the proposed approach on eight different real-world datasets covering various problem scenarios and applications. Finally, we compared the results of the introduced method with the results of using other state-of-the-art approaches in the literature. According to the results, the suggested method is capable of dynamically deciding on which feature to be acquired based on feature costs and available context in an efficient manner.

% references section

%\bibliographystyle{IEEEtran}
%\FloatBarrier
%\clearpage
%\bibliography{./bib/refs.bib}
% Generated by IEEEtran.bst, version: 1.14 (2015/08/26)

% that's all folks
\end{document}